# Fine-Tuning and Prompt Engineering of LLMs, for the Creation of Multi-Agent AI for Addressing Sustainable Protein Production Challenges

Alexander D. Kalian [1,*], Jaewook Lee [2,*], Stefan P. Johannesson [2], Lennart Otte [2], Christer Hogstrand [3], Miao Guo [2,**]

[1] Department of Nutritional Sciences, King's College London, Franklin-Wilkins Building, 150 Stamford St., London SE1 9NH, United Kingdom
[2] Department of Engineering, King's College London, Strand Campus, Strand, London WC2R 2LS, United Kingdom
[3] Department of Analytical, Environment and Forensic Sciences, King's College London, Franklin-Wilkins Building, 150 Stamford St., London SE1 9NH, United Kingdom

* Joint first-authors; equal contribution
** Corresponding author: miao.guo@kcl.ac.uk

## Abstract

The global demand for sustainable protein sources has accelerated the need for intelligent tools that can rapidly process and synthesise domain-specific scientific knowledge. In this study, we present a proof-of-concept multi-agent Artificial Intelligence (AI) framework designed to support sustainable protein production research, with an initial focus on microbial protein sources. Our Retrieval-Augmented Generation (RAG)-oriented system consists of two GPT-based LLM agents: (1) a literature search agent that retrieves relevant scientific literature on microbial protein production for a specified microbial strain, and (2) an information extraction agent that processes the retrieved content to extract relevant biological and chemical information. Two parallel methodologies, fine-tuning and prompt engineering, were explored for agent optimisation. Both methods demonstrated effectiveness at improving the performance of the information extraction agent in terms of transformer-based cosine similarity scores between obtained and ideal outputs. Mean cosine similarity scores were increased by up to 25%, while universally reaching mean scores of ≥0.89 against ideal output text. Fine-tuning overall improved the mean scores to a greater extent (consistently of ≥0.94) compared to prompt engineering, although lower statistical uncertainties were observed with the latter approach. A user interface was developed and published for enabling the use of the multi-agent AI system, alongside preliminary exploration of additional chemical safety-based search capabilities.

# 1 - Introduction

## 1.1 - Challenges in Sustainable Protein Production

The current global food system generates significant waste and carbon emissions while placing unsustainable pressure on arable land and freshwater resources [1,2]. This contributes to environmental degradation and intensifies global hunger and protein deficiency, as highlighted by Piercy et al. (2023) [1]. A promising solution lies in Microbial Protein (MP) technology, which can convert abundant carbon-rich contaminant-free side streams (often called 'waste') - such as lignocellulosic agricultural residues - into high-quality, sustainable alternative protein through fermentation [2]. MP production utilises diverse microbial strains, including yeast, fungi, bacteria, and algae, to create protein-rich biomass [1,2]. Around 80 strains have been reported for food or feed applications, with protein contents ranging from 50–80 wt% in bacteria, 60–70 wt% in microalgae, 30–50 wt% in fungi/yeasts, and 10–20 wt% in protists [1]. Despite their high protein yield, bacterial proteins face adoption barriers due to off-flavours, while fungi have a longstanding commercial record [3]. Traditional foods like tempeh and oncom, produced using *Rhizopus* or *Neurospora* species grown on agro-waste, exemplify early "waste-to-protein" pathways [3] - though upscale oncom products remain underdeveloped.

Industrial MP emerged in the 1970s with examples like *Candida*-based volatile fatty acid fermentation [4] and ICI's methanol-derived "Pruteen" [5]. However, low-cost soy protein limited market success. The most notable breakthrough is *Fusarium venenatum*-based mycoprotein, marketed as Quorn™ since 1985 [6]. Grown aerobically on glucose with added nutrients, Quorn™ offers approximately 45% protein, all essential amino acids, a healthy lipid profile, and high fibre [2, 7]. With over five billion servings to date, it shows proven health benefits and market success in regions like Europe and North America [8]. Today, MP research is increasingly focused on valorising nutrient-rich and contaminant-free side-streams using GRAS (Generally Recognised as Safe) strains [1] fed on low-grade carbon sources like glucose or xylose hydrolysates [2]. The key challenge remains achieving desirable taste, texture, and cost-effectiveness through smart formulation and process innovation [1, 2].

Despite scientific advances in sustainable protein domains, current R&D remains fragmented, manually intensive, and poorly integrated across the discovery-to-scale pipeline [3, 6, 9]. There is a critical need for intelligent systems that can reason across complex and data-intensive biological (e.g. Fig. 1), chemical, and process domains to autonomously identify, design, and optimise sustainable protein solutions.

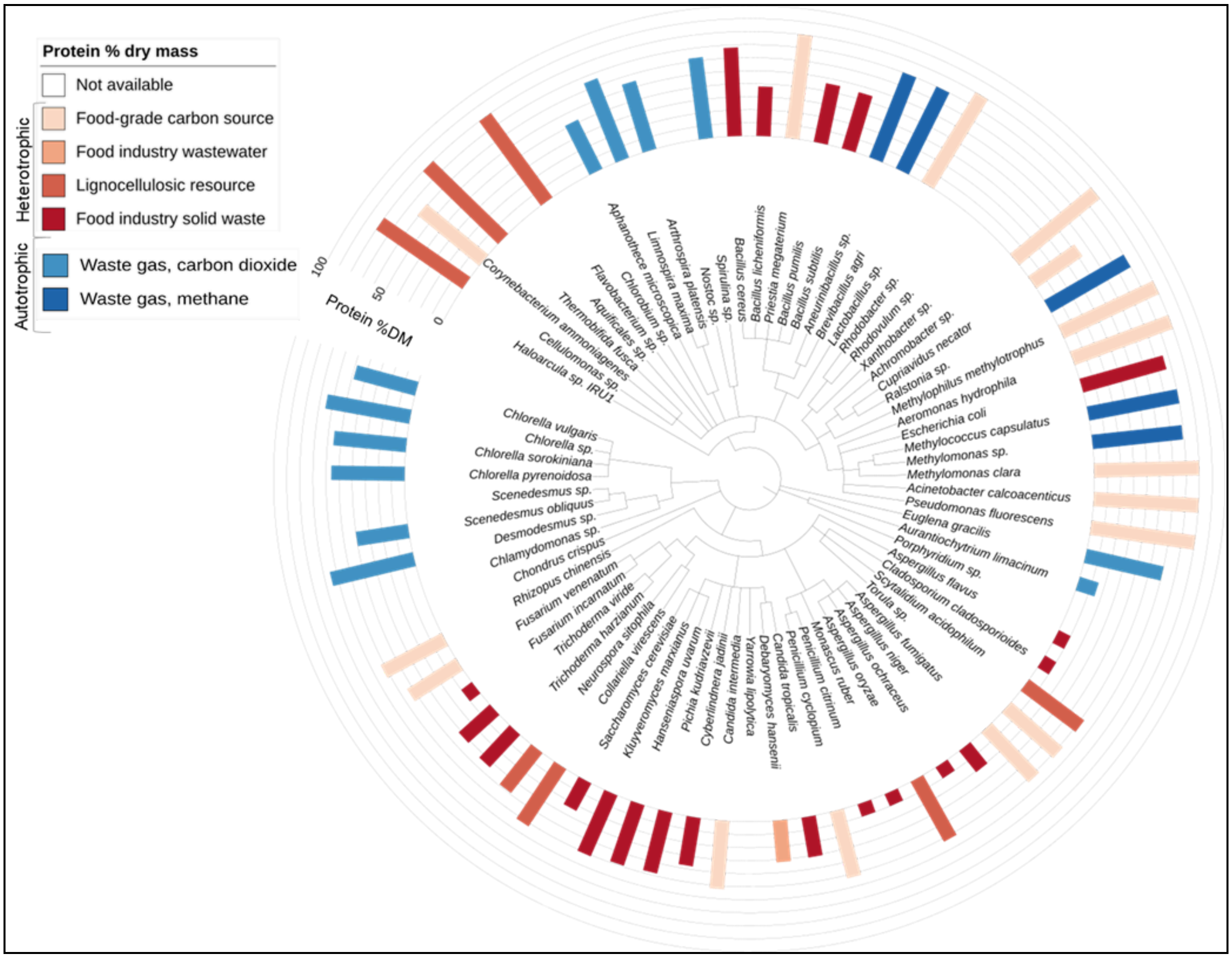

Figure 1: Taxonomic tree of reported microbial protein-producing species, extracted from Piercy et al. (2023) [1] (used with permission).

## 1.2 - Potential Use-Cases of Multi-Agent AI

Artificially intelligent Multi-Agent Systems (MASs) are designed to break down complex tasks into smaller, specialised components handled by different agents [10-12]. Each agent can focus on a specific function, such as retrieving documents, extracting structured data, reasoning about content, or generating summaries [10-12]. This allows the system as a whole to perform more complex operations than any single model could alone [10-12].

In complex and data-intensive (Fig. 1) scientific domains such as sustainable protein production, this approach is especially valuable. The research space is broad and fragmented, involving microbial biology, fermentation processes, bioeconomics, and food systems, for a large number of candidate protein-producing microbes (Fig. 1) [3, 6, 9]. Critical information, such as species characteristics, growth substrates, protein yields, and metabolic traits, is typically scattered across large volumes of academic literature [13]. Manually extracting and comparing this data is time-consuming and prone to error [13].

Multi-agent AI systems can help automate this process. For example:

- One agent can search for and retrieve papers relevant to microbial protein.
- Another can extract key information such as protein content, substrate type, and trophic mechanism.
- Additional agents could compare traits across species or identify promising candidates for specific production conditions.
- Further agents could assist in the design and optimisation of microbial protein production processes.

These agents can work sequentially or in parallel, and their outputs can be passed to human researchers or other agents for further processing [10-12]. This modularity makes the system adaptable and scalable for different research goals [10-12].

By enabling more efficient access to structured scientific knowledge, multi-agent AI has the potential to accelerate discovery and innovation in sustainable protein production. This study presents an early-stage implementation of such a system, demonstrating how multi-agent systems can support information retrieval and extraction tasks relevant to this field.

### 1.3 - Opportunities from Large Language Models

Large Language Models (LLMs), descendants of the transformer architecture first introduced in 2017, have reshaped both AI research and industry thanks to their state-of-the-art performance in natural language understanding and generation tasks [14-16]. Leveraging the transformer's scalability and parallelism, LLMs generalise more effectively than earlier architectures, leading to successful applications across a wide range of domains such as video-to-language understanding, code generation, and robotics [17-21]. Beyond this, they also demonstrate remarkable reasoning, planning, and broad world knowledge acquired during pre-training [22-25]. It is for these same reasons that LLMs have become attractive building blocks for multi-agent AI systems. By instantiating role-specific agents, one can create collaborative, modular, and adaptive frameworks that tackle complex problems through inter-agent communication; emergent collaborative behaviours have even been observed in certain LLM-based multi-agent systems [26].

A conventional approach to an LLM-based multi-agent AI system is for each agent to be powered by a language model that would then be adapted to perform a specific task. These tasks would be narrow in scope, such as retrieving relevant documents, extracting useful information, comparing results, or suggesting next steps. Because of their innate ability for natural language understanding and generation, LLM-based agents can communicate with one another simply via textual messages, while optionally embedding structured data formats for precise handoffs [27, 28]. This flexibility eases the creation of modular workflows: each agent focuses on its subtask, publishes its findings in a shared format or channel, and reads others' outputs to determine its next action. Designing or extending such systems becomes straightforward, since adding a new specialist agent often means crafting an appropriate prompt template or few-shot examples rather than writing bespoke parsing code.

LLMs are particularly well-suited for multi-agent systems, as they can handle a wide range of language-based tasks without requiring a full retraining for each role. Pretraining endows them with rich world knowledge and strong generalisation, so that with the right prompt, they can perform tasks like extracting data, summarising long passages, reasoning across inputs (including multimodal summaries), and planning or critiquing workflows [19, 29-31]. Zero-shot and few-shot learning capabilities mean that new agents or tasks can be onboarded rapidly via prompt engineering [18, 29]. This adaptability makes LLM-based multi-agent systems powerful for navigating complex domains like sustainable protein production research, where critical insights are buried in technical literature.

In broad terms, the current landscape of LLMs can be partitioned into two main categories: closed-source (proprietary) models and open-source models. Proprietary models, such as OpenAI's GPT line, Anthropic's family of Claude models, and Google's Gemini series, demonstrate strong performance across a multitude of different tasks, including tasks with multi-modal demands [16, 32, 33]. These LLMs are usually trained on large datasets using advanced forms of reinforcement learning from human feedback. As a result, these LLMs are optimised to output accurate and fluent responses, even for insufficient prompts. However, these models come with limitations: users typically are unable to access the data that the models were trained on, fine-tune them for their own purposes (except in certain limited instances), access trained model weights or run them locally. They also usually require payment, depend on external APIs (Application Programming Interfaces), and carry concerns about data privacy and long-term access [34].

On the other hand, open-source models - such as LLaMA (Meta), Mistral, Falcon, and Phi - enable researchers greater control. These models may be downloaded, inspected, modified, and arbitrarily fine-tuned on custom datasets. They can also be deployed offline or on private servers, which is important for sensitive data or reproducibility in academic research [34]. However, open-source models often trail behind their proprietary counterparts in terms of performance and reliability, especially when handling complex instructions, long documents, or multi-step tasks [35]. They furthermore tend to lack multi-modal capabilities (typically only capable of handling sequential text data), while the requirement for local use is often computationally expensive (and potentially at a greater financial cost than private subscription fees for API access to proprietary LLM models). For scientific and research-focused multi-agent systems, proprietary and open-source models each bring distinct benefits: proprietary models can speed up development and offer higher accuracy in initial phases, whereas open-source models grant greater flexibility and long-term control. The best option depends on the project's objectives, available resources, and the importance of maintaining independence from third-party APIs over time.

## 2 - Materials and Methods

### 2.1 - Research Approach and Hypothesis

In this study, we explore how LLMs can be used as the foundation for intelligent agents that assist in a RAG-based system for literature mining and data extraction [11], for the domain of sustainable microbial protein production (Fig. 2). We compare two key strategies for improving LLM-based agents: fine-tuning, where the model is retrained on domain-specific data [36], and prompt engineering, where the input is carefully crafted to guide the model's output [37]. Both approaches can utilise expert domain knowledge for improving LLM performance, albeit with fine-tuning entailing direct modification of trainable model weights [36], whereas prompt engineering may use expert domain knowledge to inform more strategic ways of constructing chains of prompts, which may guide the model to more effectively emulate expert-level outputs [37]. Our aim is to understand how best to use LLMs, via these differing approaches, to build effective, adaptable systems for supporting research in sustainable protein production.

Although numerous LLM solutions are available for modification and deployment as AI agents, this study focuses on GPT-based LLM chatbots [14], due to their industry-leading generalised performance, as well as ease of API access and external delegation of computational resources. While GPTs are closed-source black-box models [14, 24], they may serve as highly effective means for creating a multi-agent AI system that serves as a proof-of-concept.

The methodological approach entails first curating domain-specific data, before building a literature search AI agent (Fig. 2), for retrieving relevant scientific literature to a query. Following this, the suitability of different GPT models was assessed for powering the information extraction agent (Fig. 2). For this second agent, fine-tuning and prompt engineering were both explored as solutions for optimising performance.

It is hypothesised that the most current GPT versions explored will prove the most effective, as well as that both optimisation approaches for the information extraction agent will demonstrate effectiveness at performance improvement. It is further hypothesised that fine-tuning will outperform prompt engineering as an approach for optimising the information extraction agent.

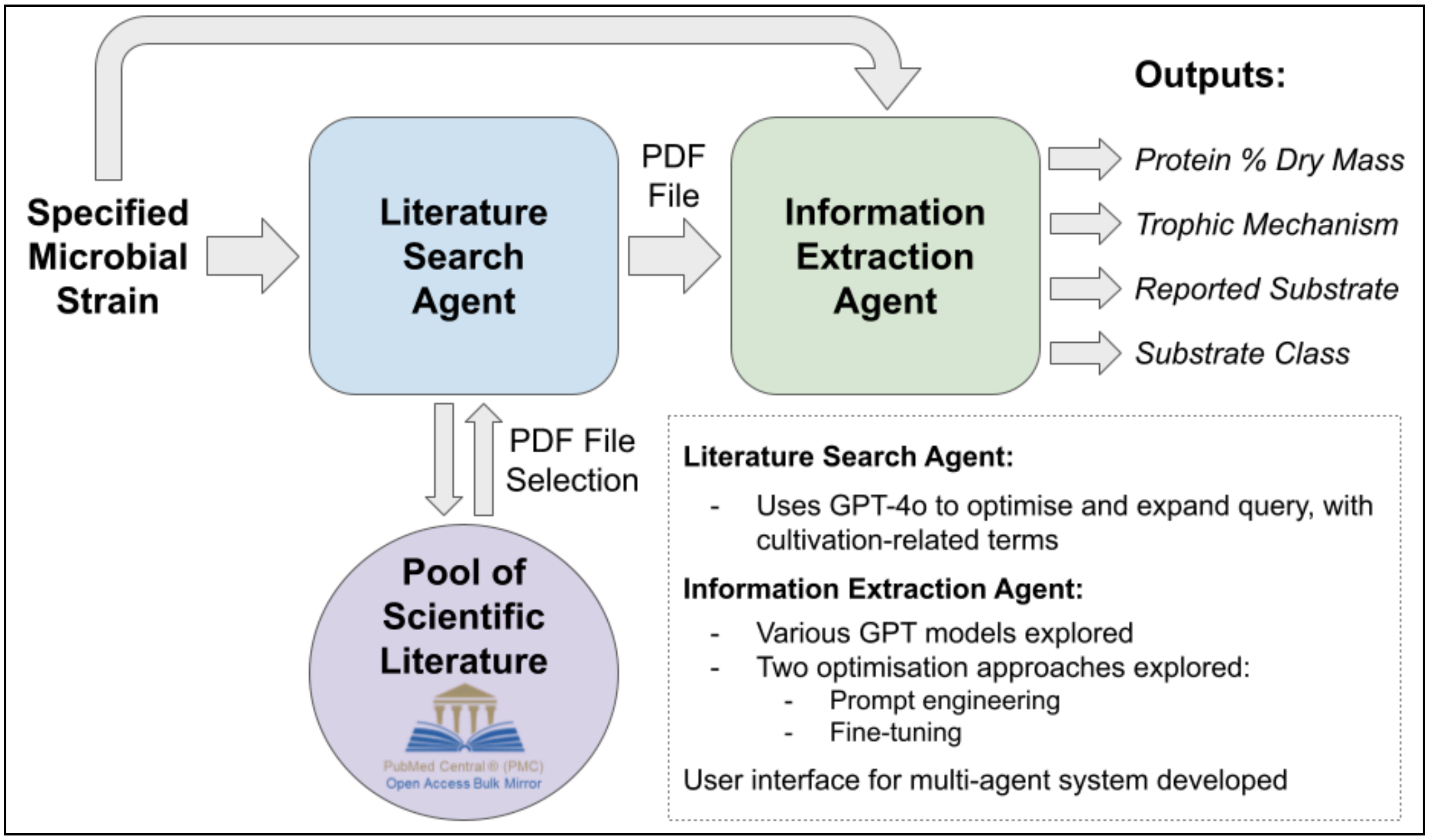

Figure 2: Schematic diagram of proposed multi-agent AI system for this study.

### 2.2 - Data Collection and Curation

An archive containing millions of open-access scientific literature was utilised, for use by the literature search agent and wider RAG system, via PubMed Central (PMC) open-access subset (Fig. 2) [38, 39].

For optimising the performance of the information extraction agent (separate from deployment as part of the wider RAG multi-agent system), more domain-specific data was leveraged. Microbial species-specific data, relevant to the proposed multi-agent system final outputs (Fig. 2), for protein% dry mass, trophic mechanism, reported substrate and substrate class, with linked studies, were obtained from the supplementary materials of Piercy et al. (2023) [1]. Text content of each associated study was then extracted by manually downloading PDF files for all papers (where available) and then parsing with the PDFMiner library in Python 3 [40]. Token count for these papers was further reduced via exclusion of the references section, as well as any detected excessive whitespace or metadata captured as text.

For cases where the token count remained excessively high (i.e. for the more limited GPT-3.5 Turbo context window), the curated PDF text data was further split into parts - a start and an end. For all such cases, this sufficiently reduced the token count.

An equal number of negative examples, of research papers without relevant specified data, were also obtained and processed, for the purposes of preventing overfitting and reducing the risk of future hallucinations in the information extraction agent [41, 42], when potentially prompted with literature text lacking sufficient information. Negative examples were equally distributed among the following 4 categories: (a) research papers about microbial protein production, but for the wrong microbial species, (b) research papers about the correct microbial species, but not relevant to microbial protein production, (c) research papers about unrelated micro-organisms, irrelevant to protein production (e.g. deep sea microbes, pathogens etc.), (d) research papers about completely unrelated topics to microbiology or protein production (e.g. astrophysics, robotics, social sciences etc.). These negative examples of papers were downloaded as PDF files and processed in the same manner as the positive examples.

Following this, the processed content of each PDF file was used to construct a respective prompt (see Section 2.4.1), for the applicable microbial strain, in a structured tabular dataset alongside corresponding ideal output expected.

### 2.3 - Creation of a Literature Search Agent

To efficiently retrieve research papers that are highly relevant to specific microbial strains and their growth conditions, we developed an integrated literature search agent that combines query optimisation, domain-specific keyword expansion, and iterative information extraction. The agent was designed to begin with a user-provided query - typically the name of a microbial species or strain, as per Fig. 2. Recognising that direct searches using only the strain name often yield papers with limited experimental detail, the agent programmatically augmented the initial query with a curated set of keywords frequently associated with microbial cultivation and physiology, such as "growth," "cultivation," "medium," "temperature," "pH," "oxygen," and "fermentation." By systematically combining the strain name with these domain-specific terms, the agent generated a series of expanded queries that targeted literature reporting detailed experimental parameters.

Each constructed query was submitted to the PubMed database through its API [38, 39], and the resulting articles were aggregated. The agent then applied a relevance scoring algorithm that evaluated the presence and frequency of both the strain name and the selected keywords within the title and abstract of each paper. Articles that surpassed a predefined relevance threshold were prioritised for further analysis. For these high-relevance articles, the agent attempted to retrieve the full text, either from open-access repositories or by parsing available abstracts.

Using the same curated set of keywords, the agent constructed multiple expanded queries by combining the original user input with various combinations of the selected keywords. For example, when the input was "Bacillus subtilis 168," the agent generated queries such as "Bacillus subtilis 168 growth conditions," "Bacillus subtilis 168 medium composition," and "Bacillus subtilis 168 temperature pH oxygen." This systematic expansion increased the likelihood of retrieving papers that reported detailed experimental parameters, as well as results relevant to the strain of interest.

### 2.4 - Creation of an Information Extraction Agent

An information extraction agent was explored to extract relevant microbial protein information from identified PDF files of scientific literature, as per the schematic overview of Fig. 2.

The information extraction agent was created via optimisation of GPT-based LLMs, with both prompt engineering and fine tuning explored as competing approaches (Fig. 3).

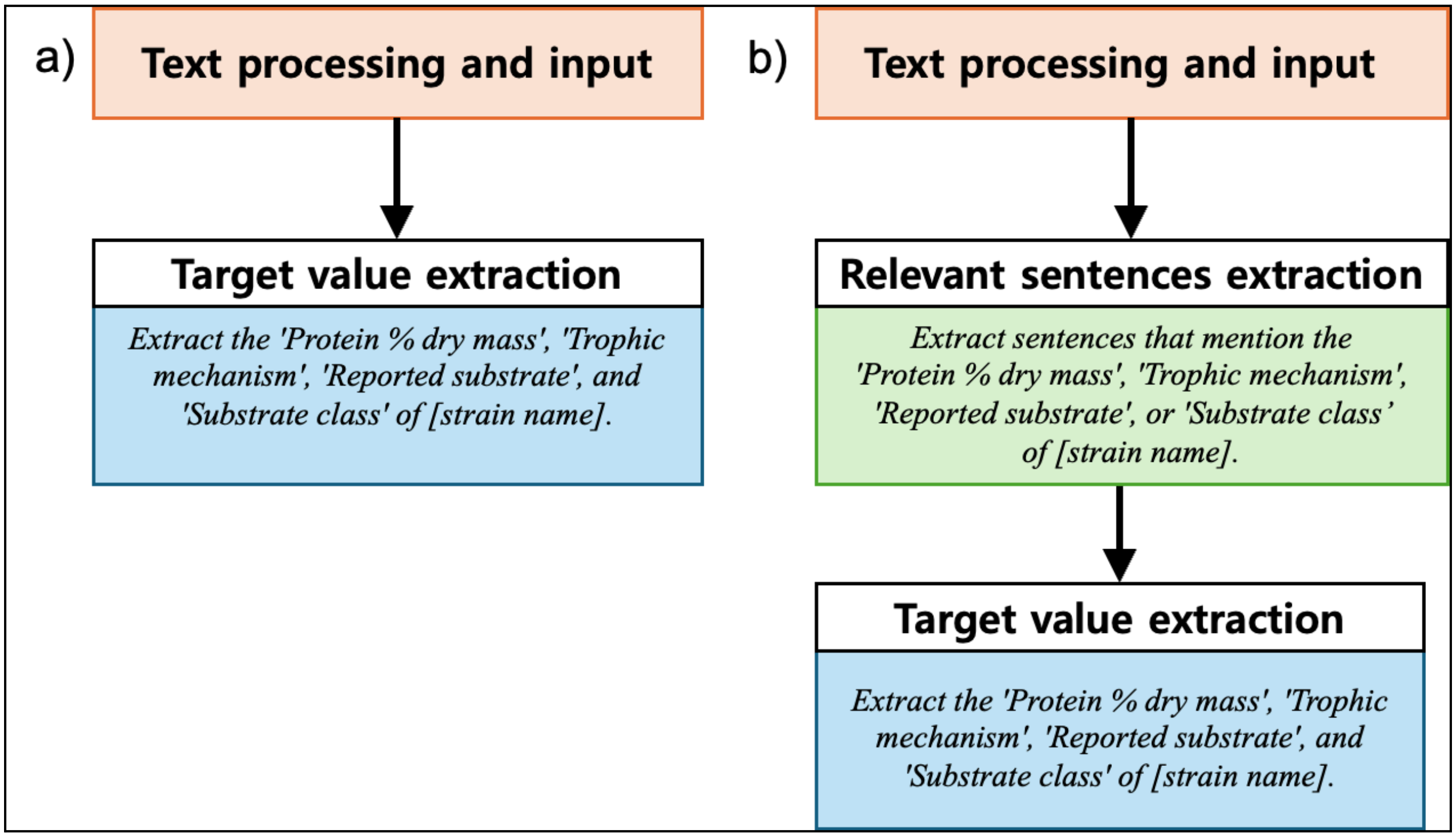

Figure 3: Schematic diagram outlining comparison between: a) the fine-tuning methodology, b) the prompt-engineering methodology; for optimising the information extraction agent.

#### 2.4.1 - Exploration of Optimal GPT Models for an Information Extraction Agent

Prior to the stages of fine-tuning and prompt engineering on a particular GPT model, in order to create an information extraction agent for processing scientific literature, it was first imperative to select an appropriate GPT model. This was done by prompting the following models: GPT-3.5 Turbo [43], GPT-4o (2024-08-06) [44] and GPT-4.1 (2025-04-14) [45] - to extract requested microbial protein production information, for a specified species and provided research study body of text. This phase of the study was carried out with the assumption that the most effective pre-existing model would be the most suitable version for subsequent fine-tuning and prompt engineering, to pursue further performance improvements.

For the rows of tabular data obtained from Section 2.1, prompts were constructed via the following format:

*“For the genus species {species_name}, please find the reported protein % dry mass, trophic mechanism, reported substrate and substrate class, from the following paper(s).*

*Please give your answer in a concise "reported protein % dry mass: [answer], trophic mechanism: [answer], reported substrate: [answer], substrate class: [answer]" format.*

*If the information cannot be found, then please respond with: "The literature provided does not contain the requested information, for the microbial species specified."*

*Find the information from the PDF text of the following paper(s):*

*{paper_text}”*

Where *{species_name}* and *{paper_text}* would be specified according to the given species and research study under consideration, respectively.

When prompting a GPT model via OpenAI’s platform, it is typical and recommended to include a “role” for the LLM to incorporate, in addition to the prompt. Although this was not a central priority in this study, the role statement “You are a helpful assistant.” was included, as it was customary and is commonly used as a *de facto* default in related studies [46, 47].

As detailed in the prompt, ideal outputs were of the following form, for positive cases:

“*reported protein % dry mass: [answer], trophic mechanism: [answer], reported substrate: [answer], substrate class: [answer]”*

Where *[answer]* would refer to the appropriate information within the paper text, as outlined in the supplementary information table from Piercy et al. (2023) [1], corresponding to the desired outputs of the multi-agent AI system of Fig. 2.

Ideal outputs were conversely of the following form, for negative cases:

“*The literature provided does not contain the requested information, for the microbial species specified.*”

For the purposes of initially gauging GPT model suitability, all curated data from the table obtained from Piercy et al. (2023) [1] was used, with no training, validation or testing data split necessary at this stage.

Temperature of a GPT model is a configurable parameter which governs the statistical rigour of next-token prediction; higher temperatures increase the tendency to more adventurous (yet potentially nonsensical or erroneous) answers [48]. While higher temperatures are desirable for generative tasks in creative fields, the information extraction task explored was entirely deterministic and hence required low temperatures (e.g. a temperature of 0.0). Nonetheless, to enable a more scientifically valid methodology, minimising assumptions for control variable values, the temperature of each GPT model was varied between 0.0 to 0.5, in increments of 0.1.

Performance was measured via comparison of obtained outputs against ideal outputs, using cosine similarity between context-inclusive embeddings from an external transformer model (Fig. 4), pre-trained on text sentence information, able to capture context-specific dependencies and similarities within otherwise seemingly differently worded strings of text (e.g. “yesterday, the cat climbed up a tree” versus “a tree was climbed by the domestic feline, the day before today”) [49]. For enabling valid comparisons and reducing dependency on a single pre-trained transformer model, 3 different models were utilised for generating text embeddings, from the open-source sentence-transformers (SBERT) library [50]: all-mpnet-base-v2, all-MiniLM-L6-v2 and sentence-t5-base.

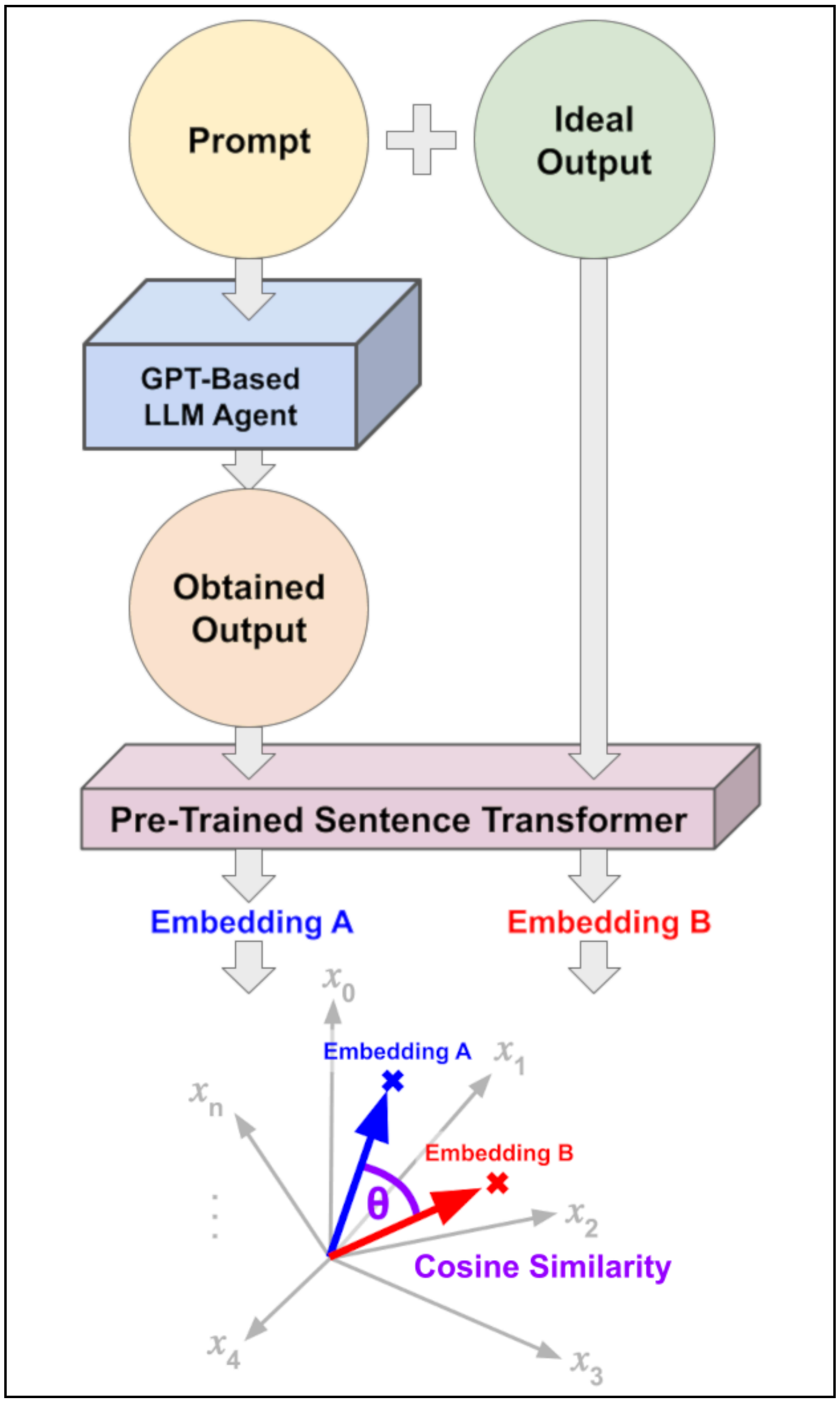

Figure 4: Schematic diagram outlining cosine similarity calculation between text embeddings of obtained versus ideal agent outputs, using an external pre-trained sentence transformer.

### 2.4.2 - Fine-Tuning for an Information Extraction Agent

Fine-tuning was used to enhance the performance of the selected GPT model to create a deployable information extraction agent. This was done using OpenAI's API over 10 epochs, with prompts and ideal answers provided as training and validation data (with testing data reserved for external validation). The split of training, validation and testing sets was assigned according to a split of 80%, 10% and 10% respectively, over the total data available, via stratified sampling. Stratified samples were according to microbial strain, such that a given microbial strain would only occur exclusively in one of the sets. Each individual datapoint, linked to a microbial strain and relevant study, also had at least one negative counterpart for that microbial strain, linked to an irrelevant study. This would enforce a balance between positive and negative cases, mitigating the risk of strain-specific biases in the final fine-tuned agents.

When fine-tuning a GPT model via OpenAI's platform, as discussed with prompting GPT models, it is typical and recommended to include a "role" for the LLM to incorporate. The previously discussed *de facto* default of "You are a helpful assistant." was used [46, 47].

OpenAI's platform automatically logs loss and accuracy scores (using functions unspecified in the documentation), over the training and validation sets. Furthermore, fine-tuned model checkpoints are saved over each of the final 3 epochs - which provides 4 available model states, when also considering the base GPT model at epoch 0. Calculation of cosine similarities between obtained and ideal outputs over the external testing set, as per Fig. 4, was used to monitor performance over the fine-tuning process.

### 2.4.3 - Prompt Engineering for an Information Extraction Agent

| Stage | Goal | Prompt design strategy | Expected output |
|---|---|---|---|
| Context harvesting | Enumerate every passage that may contain the target facts while preserving full experimental context | "Exploratory-recall" prompt that assigns a microbiology-expert role, issues exhaustive search instructions, requires verbatim reproduction of quantitative values, and forbids lossy summarisation so that adjacent sentences, figure captions, and table headers are retained | Four field-specific text blocks, each containing contextual sentences or the sentinel string *No relevant content found* |
| Canonical value extraction | Compress noisy candidate text into a single canonical value for each field | "Constrained-generation" prompt that supplies positive/negative schema examples, enforces schema-only output, encodes precedence rules (strain-specific > genus-level). | One scalar or short string per field, or *nan* if evidence is absent |

Table 1: Two-stage prompt architecture implemented in the extraction agent.

Prompt engineering is treated here not as an *ad hoc* scripting convenience but as a reproducible methodological instrument whose parameters can be inspected and interrogated. Recent surveys emphasise that the behaviour of LLMs can be “programmed” through carefully structured natural-language instructions, obviating additional fine-tuning yet yielding competitive task performance [51, 52]. Guided by that premise, we decomposed the task into the complementary stages summarised in Table 1, thereby balancing the high recall demanded by scientific evidence gathering with the precision required for quantitative synthesis. This decomposition echoes emerging work in generative relation-extraction pipelines for specialised biomedical domains, where an initial summarisation phase precedes value normalisation [53, 54].

The first-stage prompt conditions the model with an explicit disciplinary identity and insists on exhaustive retrieval of context, a technique shown to enhance domain alignment and reduce omission of low-frequency terminology in scientific documents. Critically, all retrieval instructions are cast as mandatory clauses within a single paragraph to exploit the model’s tendency to obey syntactically proximal constraints. Because LLMs are empirically prone to hallucination when forced to answer despite insufficient evidence, we insert an early-exit gate that halts the chain when the harvested context is demonstrably sparse, an approach recommended in recent hallucination-mitigation taxonomies [41, 42]. The second-stage prompt adopts the opposite stance: it constrains output space to a one-token or short-phrase schema, reinforced by positive and negative examples. This “schema anchoring” technique forces the model to represent uncertainty explicitly using a special placeholder—such as a “NaN” (Not a Number) sentinel—rather than producing speculative answers. This aligns with recent proposals advocating for clearer uncertainty signalling in LLM outputs [55].
Performance was therefore assessed on precisely the same benchmark test set and according to the identical evaluation metrics that had been applied to the fine-tuned model, ensuring strict methodological parity and permitting a direct comparison between the prompt-engineered and fine-tuned approaches.

## 2.5 - User interface development and toxin synthesis pathways

To facilitate intuitive and efficient interaction with the microbial strain research agent, we developed a web-based user interface using Streamlit [56], an open-source Python framework designed for rapid prototyping of data-driven applications. This interface is structured to guide users through the end-to-end process of configuring, executing, and interpreting automated literature analyses related to microbial growth conditions.

As shown in Fig. 5a, users begin by entering the species name and selecting the maximum number of papers to analyse using a simple configuration panel. Upon initiating the analysis, the system automatically retrieves and processes relevant literature. The core results are displayed in an organised dashboard Fig. 5b, which summarises key experimental findings such as protein content (% dry mass), trophic mechanism, reported substrates, and substrate classes. Each value is displayed within a colour-coded metric card for quick assessment, with additional contextual information such as the number of supporting papers. To assess potential biological risks, a dedicated toxicity analysis is performed (Fig. 5c), which visualises the mutagenicity potential of compounds identified in the biological pathway. Furthermore, the GPT Search Optimisation History panel (Fig. 5d) offers traceability by listing all queries used, relevant results retrieved, and key terms identified. It includes logs of which content was successfully extracted and which failed, providing feedback for iterative improvement.

The interface follows a progressive, stepwise workflow encompassing initialisation, literature search, and analysis stages. Real-time progress indicators and status messages provide transparency regarding the system's current state and ongoing operations. Detailed logs of all user actions, query formulations, agent decisions, and outcomes are maintained and viewable in dedicated expandable tabs.

Overall, the streamlined single-page design and custom styling enhance readability and usability, enabling both domain experts and non-specialists to interact with the system effortlessly. By leveraging Streamlit's interactivity and responsiveness, the platform ensures transparency, ease of use, and reproducibility in automated literature analysis.

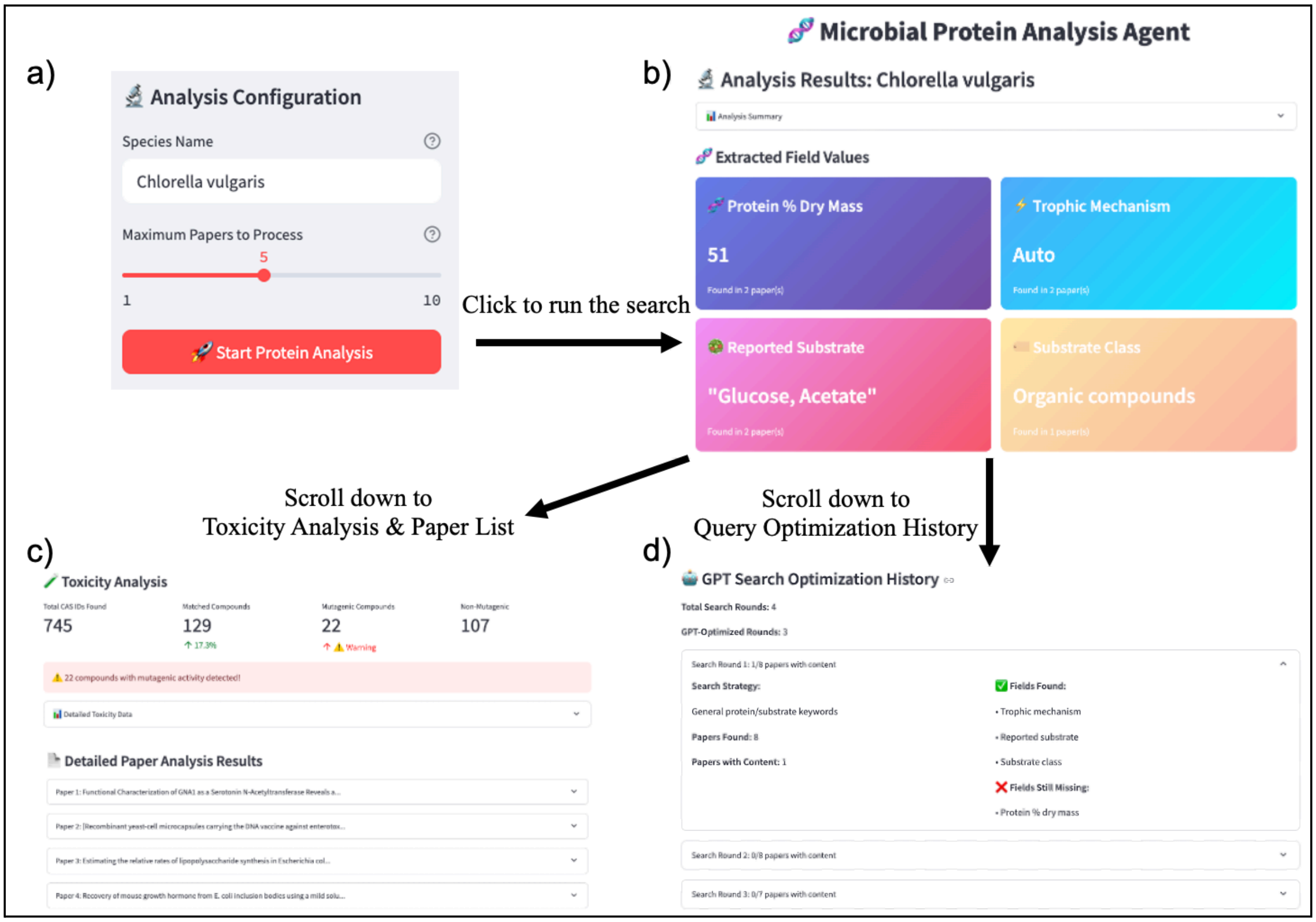

Figure 5: Overview of user interface developed for utilising the multi-agent AI system.

### 2.6 - Biological Synthesis Pathways - Toxicity Screening

Synthesis pathways are part of the metabolic network of an organism, which can be screened for potentially harmful intermediates. With the advancement of multi-omics techniques, vast amounts of the genome, proteome and metabolome are measured and used to infer pathways (synthesis as well as other biological functions). Inferred and partially experimentally validated pathways are stored in Pathway Databases (PDBs) such as KEGG [57], the BioCyc [58] collection and Reactome [59]. In this work, we use BioCyc to identify the compounds that comprise the metabolome of an organism. BioCyc contains ~20,000 organism-specific databases and is thus suitable to screen the metabolome of a large number of organisms. Using the BioCyc web services, we can query the compounds of a given organism database and cross-check against toxicity (mutagenicity) datasets [60] to avoid toxic metabolites. At first, we query for all compounds involved in the metabolic network of the specified organism. Secondly, we match the CAS numbers of the compounds returned by the database with the toxicity dataset and highlight toxic compounds.

## 3 - Results and Discussion

### 3.1 - Optimal GPT Model for Information Extraction Agent

As shown in Fig. 6, the exact cosine similarity scores varied depending on the external pre-trained sentence transformer used to generate embeddings; however, the relative trends remained closely consistent.

The sentence-t5-base transformer (which was the most expressive of the transformer models used), of Fig. 6c, appeared to evaluate the obtained outputs with stronger cosine similarities to ideal outputs (all >0.85). Conversely, the all-MiniLM-l6-V2 sentence transformer (which was the least expressive model), of Fig. 6b, evaluated certain outputs as closer to ~0.6 in cosine similarity to corresponding ideal outputs. The all-mpnet-base-v2 sentence transformer, of Fig. 6a, served as a middle ground between these two models, with cosine similarities ranging between 0.7-0.85.

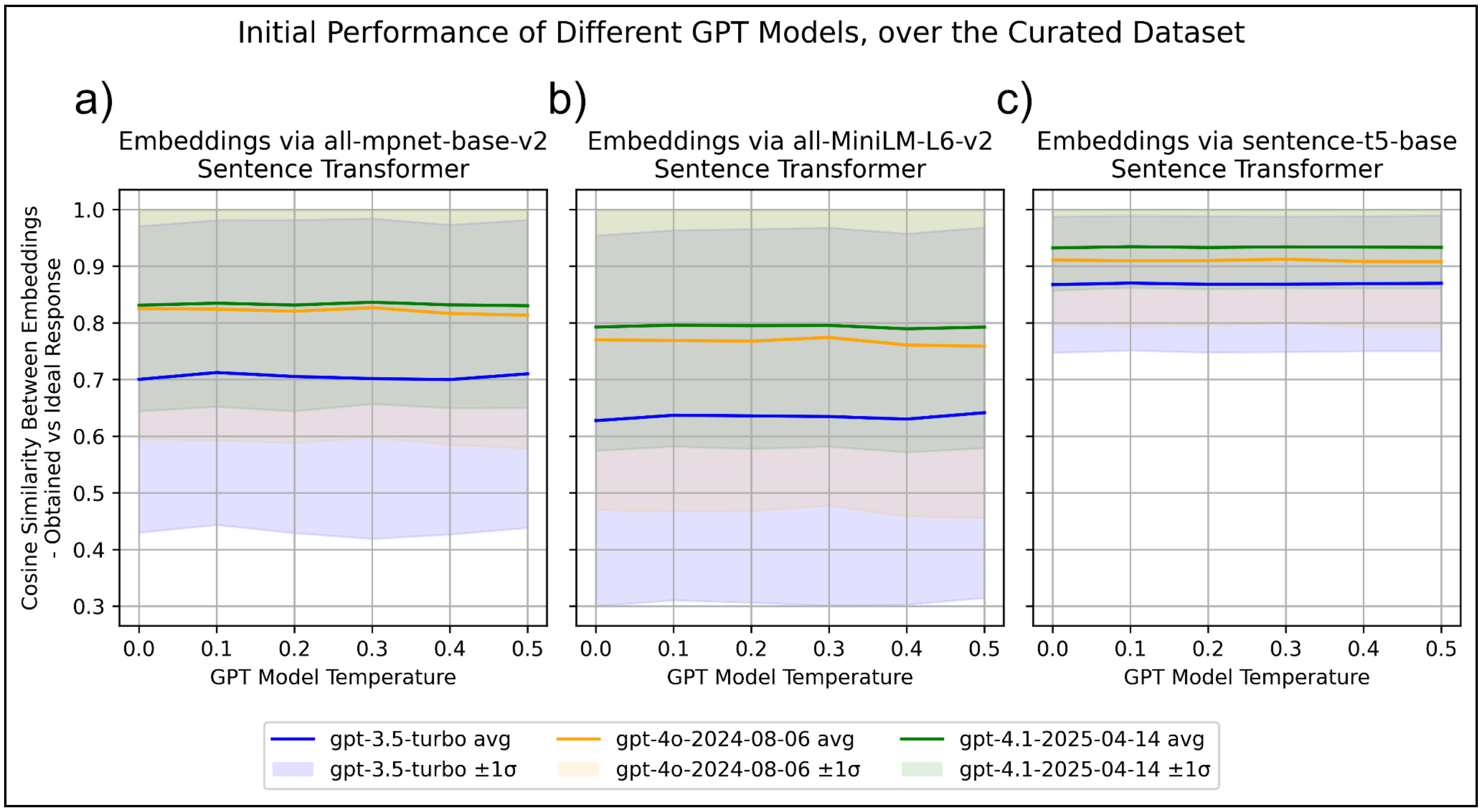

Figure 6: Performance of the different explored GPT models, for gauging suitability for the information extraction agent task, over the whole curated dataset; varied by GPT model temperature and by sentence transformer model used for evaluation.

Regardless of the sentence transformer model used for accuracy evaluation, consistent trends of GPT-4.1 (2025-04-14) being the strongest pre-existing model held across Fig. 6, hence marking it as the most suitable model for subsequent fine-tuning and prompt engineering efforts. This was closely followed by GPT-4o (2024-08-06), while GPT-3.5 Turbo significantly underperformed both GPT-4 based models in performance (which is to be expected, given that it is a legacy model in comparison).

A significant degree of statistical overlap (computed via standard deviation values) was found between GPT model performances in Fig. 6, hence indicating high degrees of variability of performance, over specific prompts. The average performance values however remained closely consistent across different trialed GPT model temperature values, hence indicating that the statistical error overlaps were representative of reproducible distributions in performance across the curated dataset, while also representing a null finding for any impact of model temperature on model performance, within the explored range of values.

Due to the insignificant impact of GPT model temperature on performance, as apparent in Fig. 6, temperature values were set to 0.0 (to ensure maximally deterministic predispositions of the GPT models), for subsequent analyses.

### 3.2 - Analysis of Fine-Tuned Information Extraction Agent

The results displayed in Fig. 7 demonstrate that the fine-tuning of GPT-4.1 (2025-04-14) was a quantifiable success, improving performance throughout the fine-tuning process (as far as the available model checkpoints for evaluation indicate), across the testing set.

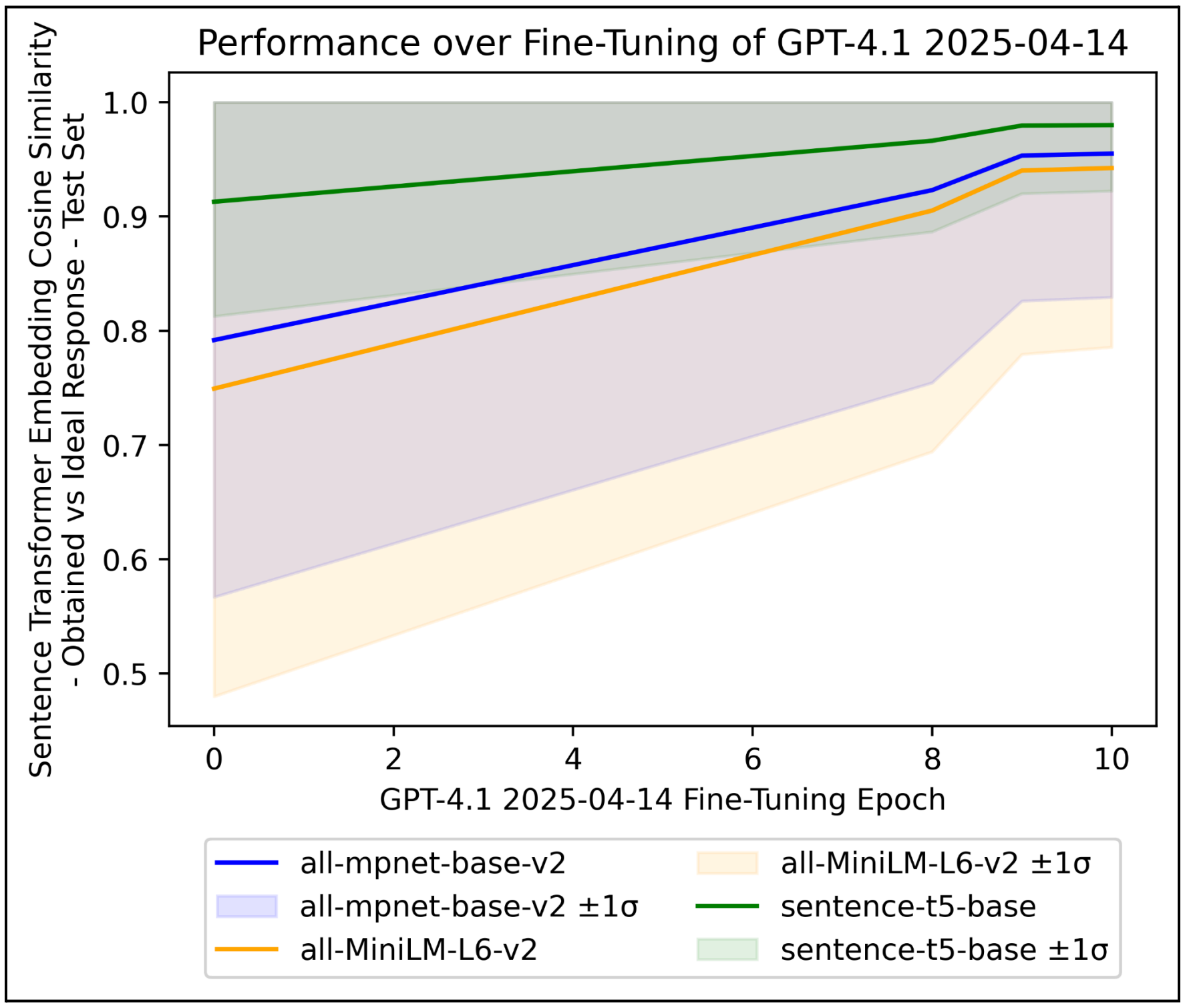

Figure 7: Fine-tuning results over the external testing set, for optimising performance of the information extraction agent.

While specific cosine similarity scores differed by transformer model used for analysis in Fig. 7, the overall trends obtained were identical, across the available model checkpoints. Interestingly, the final cosine similarities (from different external sentence transformers) converged in closer mutual proximity, alongside a narrowing of uncertainty ranges. This may indicate a convergence towards an objective optimum, in optimising the agent task of information extraction from scientific literature; the different sentence transformer evaluation models may have offered diversified insights into the context-based similarities and differences between the obtained and ideal outputs, with all closely satisfied by a fine-tuned LLM agent able to more closely output ideal responses, even over external testing data.

Clearer insight into the improvements of the final fine-tuned model, against the base model case, are outlined in Fig. 8. As per Fig. 7, differences in exact cosine similarity scores varied between different sentence transformers used for evaluation, however the trend of improved performance from fine-tuning was unanimously demonstrated.

While an extent of overlap in statistical uncertainty is present in Fig. 7 and Fig. 8, across all evaluations on the testing set, clear trends of higher average cosine similarity scores, coupled with narrower statistical uncertainty ranges are apparent; hence a positive shifting of the cosine similarity score distributions, across the external testing set, indicative of successful fine-tuning.

The improvements in average cosine similarity scores in Fig. 7 and Fig. 8 demonstrate effectiveness of the fine-tuning stage. Average cosine similarity improved from 0.79 to 0.96 (2 s.f.) when evaluated via the all-mpnet-base-v2 sentence transformer - representing a 22% (2 s.f.) increase in performance. Improvements from 0.75 to 0.94 (2.s.f), as well as 0.91 to 0.98 (2 s.f.), via evaluation using the all-MiniLM-l6-V2 and sentence-t5-base transformers, were reached, respectively representing improvements of 25% and 8%.

As all external transformer models independently produced evaluated cosine similarity scores of ≥0.94, it may hence be deduced that the fine-tuning of the information extraction agent was both unequivocally successful and generalised well to external testing data.

### 3.3 - Analysis of Prompt Engineered Information Extraction Agent

Since prompting, unlike fine-tuning, does not encode the answer text into the model parameters unless explicitly embedded in the prompt, the risk of inadvertent data leakage is inherently low. Nevertheless, to ensure a fair comparison with fine-tuning approaches, the prompt-engineered information-extraction agent was evaluated using the same train–validation–test partitions as the fine-tuned models, thereby eliminating any confounding effects from divergent data splits. Based on the results presented in Section 3.1, we employed GPT-4.1 (2025-04-14) for this evaluation, as it demonstrated the best performance among the GPT series. For each test instance we compared the model's predicted extraction with an "ideal" reference using three sentence-embedding models and reported cosine similarity. As illustrated in Fig. 8, introducing a two-stage prompt - in which the model first identifies semantically salient candidate sentences and then extracts the target value from that subset - raised the mean similarities across all three embeddings from 0.79, 0.78, and 0.91 (2 s.f.), to 0.92, 0.89, and 0.96 (2 s.f.), corresponding to absolute gains of 0.13, 0.11, and 0.05 (representing respective percentage improvements of 16%, 14% and 5%). This improvement was accompanied by a pronounced reduction in uncertainty, with the standard deviations decreasing from 0.18, 0.15, and 0.08 (2 s.f.) in the baseline to 0.075, 0.093, and 0.033 (2 s.f.) after prompt engineering. These results demonstrate that the structured two-stage prompt not only delivers unequivocal performance gains but also attenuates the model's prediction variance, thereby mitigating one of the principal limitations of LLMs in information-extraction tasks.

### 3.4 - Comparison of Fine Tuning and Prompt Engineering

Both optimisation routes were evaluated on identical train–validation–test partitions and scored with three independent sentence-embedding models. The GPT-4.1 (2025-04-14) fine-tuned model retained the previously reported mean cosine similarities (with use of standard deviation as statistical uncertainty) of 0.96 ± 0.13, 0.94 ± 0.16 and 0.98 ± 0.06, when assessed with

all-mpnet-base-v2, all-MiniLM-L6-v2 and sentence-t5-base sentence transformer models, respectively (see Fig. 7 and Fig. 8). Re-analysis of the two-stage prompt-engineered system with the final evaluation protocol yields refined estimates: a mean similarity of 0.92 ± 0.075 via all-mpnet-base-v2, 0.89 ± 0.09 via all-MiniLM-L6-v2, and 0.96 ± 0.03 via sentence-t5-base (Fig. 8). These figures place the absolute advantage of fine-tuning at 0.04, 0.05 and 0.02 respectively on the three metrics, confirming a consistent head-room of three to five percentage points except on sentence-t5-base, where the margin narrows to roughly two points.

The mean scores and standard deviations of Fig. 8 overall suggest that fine-tuning delivers superior mean performance but with higher statistical uncertainty, whereas prompt engineering offers a lighter-weight and lower statistical uncertainty (albeit lower performance) alternative. Although the mean scores of fine-tuning are consistently evaluated as higher than for prompt engineering, there is significant statistical uncertainty overlap, hence suggesting an indicative finding only.

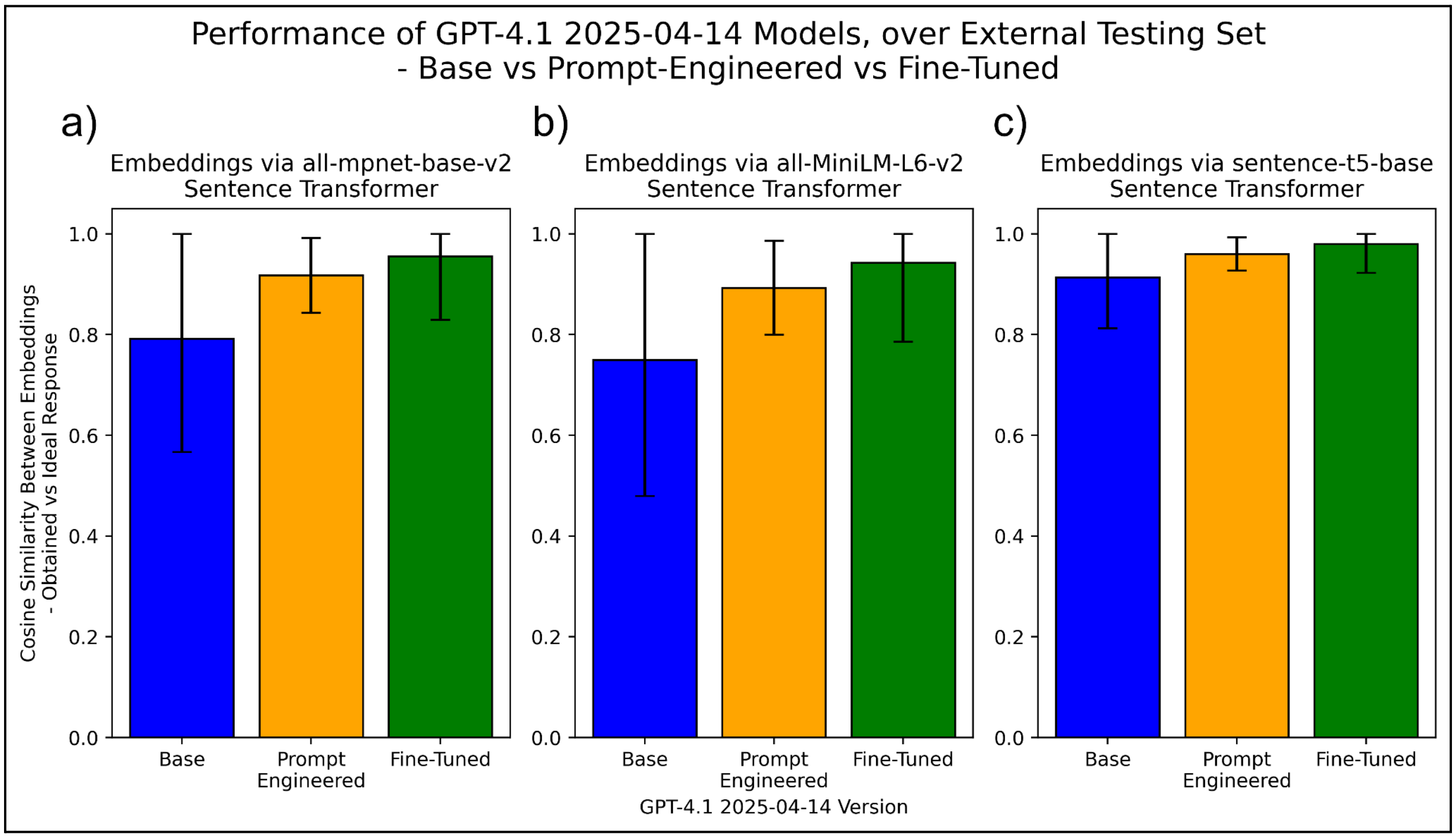

Figure 8: Bar chart comparing performance between the base, prompt engineered and fine-tuned gpt-4.1 2025-04-14 versions, for the information extraction agent task.

While the fine-tuning approach appeared to yield greater mean performance scores, compared to the prompt engineering approach, related research suggests that fine-tuning may worsen calibration of an LLM, compared to prompt-engineering, increasing confidence of incorrect answers [61]. While our study may have mitigated this risk, via use of a balanced dataset containing negative examples of unrelated literature, as well as quantifying final performance against an external testing set, the fine-tuned LLMs may nonetheless be less generalisable in their trained states, potentially underperforming over novel data points that stray further from the entire dataset used. While only a speculative limitation of the fine-tuned LLMs, for the information extraction task explored, this potential limitation should be acknowledged, in the scope of comparing fine tuning and prompt engineering as methodologies for improving LLM-based agent performance.

### 3.5 - Future Outlook

Drawing on the success of this preliminary study, for optimising LLMs into AI agents, for a multi-agent AI framework, subsequent research may provide benefit through various adjacent expansions. Examples may include more rigorous fine-tuning and prompt engineering regimes, with more extensive data, detailed information requested from associated studies and increased number of fine-tuning epochs. Additionally, the utilised performance metric of cosine similarity between sentence transformer embeddings, while effective at a broad overview of LLM output performance, also holds risks as a potentially over-generalised and superficial performance metric, for optimising against a use-case that requires precise numerical and technical information. Future expansions of this research may hence explore isolation of individual categories of information from outputs (e.g. protein % dry mass) and then individually quantifying error on these values against true values. For categorical output information, such as trophic mechanism, a more isolated cosine similarity of ideal versus obtained output text embeddings may take place, solely on this information (and perhaps utilising better suited transformer models for producing meaningful embeddings of these briefer text strings).

Furthermore, as GPT models are proprietary and closed-source [14, 24], future expansions of this project would benefit from use of open-source LLMs such as Llama 2 [62] or DeepSeek-V3 [63], with optimised model states readily available and able to be compared against other fine-tuned open-source deviations of these LLMs in related studies.

The loss function and accuracy metric, used by the OpenAI platform, over the fine-tuning stage for the training and validation data, were not specified in OpenAI's documentation. Use of an open-source LLM in future expansions may hence also enable more transparent use of loss functions and accuracy metrics, evenly applicable across the training, validation and testing data, as well as with fine-tuned states at each epoch readily available.

Future expansions may also create other AI agents to complement the multi-agent AI system pipeline of Fig. 2, building on this preliminary proof-of-concept, with performance metrics evaluated over the whole interconnected multi-agent system, in addition to on an individual fine-tuning and prompt engineering level. Given the high degree of success of both fine-tuning and prompt engineering of agents, observed in this study, a combined methodology which simultaneously utilises both approaches, may be explored in future expansions, to garner even further benefit.

## 4 - Conclusions

From our range of results and analyses, it may be concluded that both fine-tuning and prompt engineering serve as highly effective approaches for optimising performance of LLM-based AI agents, for the purposes of extracting information from literature relevant to sustainable protein production. Furthermore, a diverse range of independent sentence transformer models used to compute cosine similarity based performance metrics, which demonstrated highly compatible agreement in the overall trends uncovered, hence further affirming the reliability of the results. Because the two explored LLM-based agent optimisation approaches of fine-tuning and prompt engineering are able to be applied complementary for a given agent, we posit that a dual-track optimisation strategy that first designs task-specific prompts and then applies lightweight fine-tuning would yield an even stronger system without prohibitive cost.

This preliminary study further demonstrated effective creation of a research paper obtainment agent, an intuitive browser-based UI and an automated toxicity-analysis module. Each component is surfaced through well-documented APIs and packaged as a plug-and-play microservice. This deliberately modular architecture will enable forthcoming Life-Cycle Assessment (LCA) evaluators, process-analysis dashboards, and other domain-specific modules to integrate seamlessly with the existing pipeline, thereby accelerating end-to-end research workflows. Future work will therefore explore the dual-track optimisation strategy while enlarging the benchmark corpus, extending the framework to open-source foundation models, and measuring end-to-end impact in realistic discovery scenarios. Ultimately, we aim to cultivate an ecosystem of interoperable agents that accelerates innovation in sustainable protein production.

## Acknowledgements

This research has been generously funded and supported by the Bezos Earth Fund, through a Phase I grant under the AI Grand Challenge for Climate and Nature.

ADK is supported by grants from the Biotechnology and Biological Sciences Research Council [grant number BB/T008709/1] and the Food Standards Agency [Agency Project FS900120].

The authors would like to acknowledge the financial support from the King's College London Net Zero Centre Ph.D. Scholarship scheme.

## Author Contributions

ADK, JL and MG conceptualised the study. MG and CH supervised the study. ADK, JL and LO carried out computational experiments for the study. ADK and JL produced analyses of the results. ADK, JL, SFJ, LO and MG wrote the manuscript. ADK and JL produced diagrams for the manuscript. ADK, JL, SFJ, LO, CH and MG reviewed and edited the manuscript.

## Supplementary Information

Python scripts, other software developed and data used in this study, shall be made available via the following GitHub repository: https://github.com/MGuo-Lab/LLM4SustainableProtein